\pdfoutput=1

\documentclass[11pt]{article}

\usepackage{pdfpages}
\usepackage{float}
\usepackage{caption}

\usepackage[]{acl}

\usepackage{times}
\usepackage{latexsym}
\usepackage{xspace}

\usepackage[T1]{fontenc}

\usepackage[utf8]{inputenc}

\usepackage{microtype}

\usepackage{inconsolata}

\usepackage{xcolor}
\definecolor{customwhite}{HTML}{FFFEFF}

\usepackage{amsmath}
\usepackage{amsthm}
\usepackage{amssymb}
\usepackage{color}
\usepackage{mathtools}
\usepackage{makecell}
\usepackage{todonotes}
\usepackage{diagbox}
\usepackage{tablefootnote}
\usepackage{layouts}
\usepackage{nicefrac}
\usepackage{multirow}
\usepackage{booktabs}
\usepackage{graphicx}
\usepackage{adjustbox}
\usepackage{tabularx}

\usepackage{soul}

\usepackage{algorithm}
\usepackage[noend]{algpseudocode}

    \algnewcommand\algorithmicto{\textbf{to}}
    \algnewcommand\To{\algorithmicto{} }
    \algnewcommand\algorithmicswitch{\textbf{switch}}
    \algnewcommand\algorithmiccase{\textbf{case}}
    \algdef{SE}[SWITCH]{Switch}{EndSwitch}[1]{\algorithmicswitch\ #1\ \algorithmicdo}{\algorithmicend\ \algorithmicswitch}%
    \algdef{SE}[CASE]{Case}{EndCase}[1]{\algorithmiccase\ #1}{\algorithmicend\ \algorithmiccase}%
    \algtext*{EndSwitch}%
    \algtext*{EndCase}%

\DeclareMathOperator*{\argmax}{arg\,max}
\DeclareMathOperator*{\argmin}{arg\,min}

\DeclareMathOperator{\E}{\mathbb{E}}

\let\baraccent=\= %
\renewcommand{\=}[1]{\stackrel{#1}{=}} %

\providecommand{\R}{\mathbb{R}}

\providecommand{\cL}{\mathcal{L}}

\providecommand{\cX}{\mathcal{X}}

\renewcommand{\P}{\mathbb{P}}
\newcommand{\PLLM}{\mathbb{P}_{\mathsf{LLM}}}

\mathchardef\mhyphen="2D %

\definecolor{darkblue}{rgb}{0.0, 0.0, 0.55}

\newcommand{\Unif}{\mathrm{Unif}}

\providecommand{\cH}{\mathcal{H}}

\newcommand{\interior}[1]{%
  {\kern0pt#1}^{\mathrm{o}}%
}

\newcommand{\smax}{\mathrm{smax}}

\def\bW{{\boldsymbol W}}
\def\bX{{\boldsymbol X}}

\def\bZ{{\boldsymbol Z}}

\def\ba{{\boldsymbol a}}
\def\bb{{\boldsymbol b}}

\def\bd{{\boldsymbol d}}

\def\bp{{\boldsymbol p}}

\def\bv{{\boldsymbol v}}

\def\bx{{\boldsymbol x}}

\newcommand{\tba}{\tilde{\ba}}
\newcommand{\tbb}{\tilde{\bb}}

\newcommand{\tbp}{\tilde{\bp}}

\newcommand{\KL}{\mathrm{KL}}

\newcommand{\hbp}{\hat{\bp}}

\newtheorem{definition}{Definition}

\usepackage[most]{tcolorbox}
\usepackage{listings}
\usepackage{xcolor}

\title{
Prompts have evil twins

}

\author{Rimon Melamed \\
  GWU \\
  \texttt{rmelamed@gwu.edu} \\\And
  Lucas H. McCabe \\
  GWU and LMI \\
  \texttt{lucasmccabe@gwu.edu} \\\And
  Tanay Wakhare \\
  MIT \\
  \texttt{twakhare@mit.edu} \\\AND
  Yejin Kim \\
  GWU \\
  \texttt{yejinjenny@gwu.edu} \\\And
  H. Howie Huang \\
  GWU \\
  \texttt{howie@gwu.edu} \\\And
  Enric Boix-Adsera \\
  MIT \\
  \texttt{eboix@mit.edu} \\}

\begin{document}
\pagenumbering{arabic}
\thispagestyle{plain}
\pagestyle{plain}

\maketitle

\begin{abstract}
We discover that many natural-language prompts can be replaced by corresponding prompts that are unintelligible to humans but that provably elicit similar behavior in language models. We call these prompts ``evil twins'' because they are obfuscated and uninterpretable (evil), but at the same time mimic the functionality of the original natural-language prompts (twins). Remarkably, evil twins transfer between models. We find these prompts by solving a maximum-likelihood problem which has applications of independent interest.\footnote{Our code and data is available at \url{https://github.com/rimon15/evil_twins}}.
\end{abstract}

\section{Introduction}
\label{sec:introduction}

Large Language Models (LLMs) are rapidly improving across a wide range of tasks~\citep{openai2023gpt,touvron2023llama,touvron2023llama2,jiang2023mistral,bubeck2023sparks}. LLMs are typically instruction-tuned~\citep{ouyang2022training} to accept user queries as prompts, and these prompts have become the primary interface for interacting with these models. Nevertheless, many basic questions on how models parse prompts remain largely open. In this paper, we examine the question:
\begin{center}
\textit{Do language model prompts have to be understandable by humans in order to elicit desired behavior?} 
\end{center}
This question has far-reaching relevance, both to engineering prompts in order to maximize performance, and for safety (e.g., uninterpretable prompts could be used to bypass safety filters and induce malicious behaviors in language models); see discussion in Section~\ref{sec:related}.

\subsection{Our contributions}
The main contribution of this paper is to build negative evidence towards the above question. We show that natural-language prompts can often be replaced by prompts that are unintelligible to humans, but that cause the model to behave \textit{functionally} similarly to the original natural-language prompt. In more detail:

\begin{figure*}[htbp]
    \centering
    \includegraphics[clip, trim={2.75cm 18.1cm 2.75cm 2.7cm}, width=1.0\textwidth]{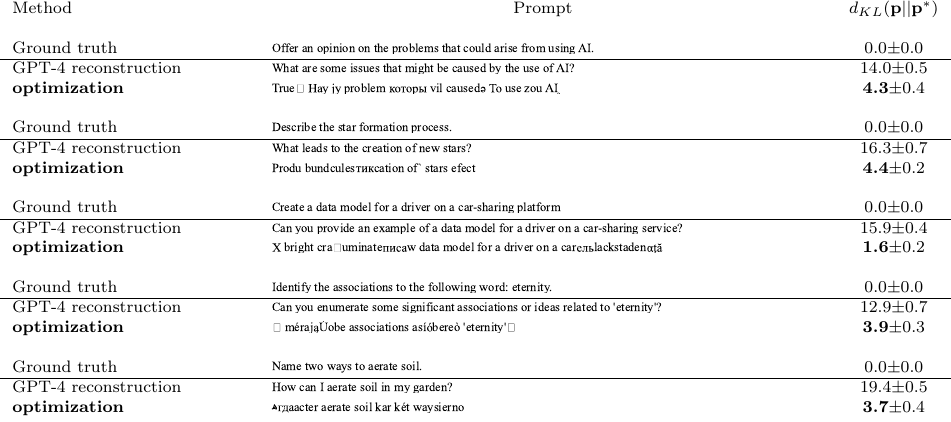}
    \caption{Five examples of ground truth prompts $\bp^*$ and corresponding ``evil twins'' $\bp$. Each evil twin is found by solving the maximum-likelihood problem \eqref{eq:mle-def-intro} on 100 documents generated from the ground truth prompt. We compare the evil twins to a baseline created by asking GPT-4 to generate a prompt that could have created the 100 documents. Surprisingly, the optimized prompts, although incoherent, are more functionally similar to the ground truth prompt (lower KL divergence) than the GPT-4 reconstruction. Details are in Section~\ref{sec:methods-comparison}. Figure~\ref{fig:full-kl-results} in the appendix contains a full table of results.}
    \label{fig:prompt-optimized-examples}
\end{figure*}

\paragraph{Functional similarity between prompts} First, we propose a quantitative measure of functional similarity between two prompts $\bp$ and $\bp^*$, by viewing them as inducing distributions $\PLLM(\cdot | \bp)$ and $\PLLM(\cdot | \bp^*)$ over outputs when fed into a language model. The two prompts are functionally similar if these distributions are similar, which we measure through the Kullback-Leibler divergence (KL):
\begin{align}\label{eq:kl-defn-llm}
d_{KL}(\bp^* \| \bp) := \KL(\PLLM(\cdot | \bp^*) \| \PLLM(\cdot | \bp)).
\end{align}
The KL divergence is an information-theoretic measure of the distance between two distributions, which is zero if and only if the two distributions are identical \citep{cover1991entropy}.

\paragraph{Finding prompts with similar functionality}

Given a ground-truth prompt $\bp^*$, we seek to find a functionally similar prompt $\bp$. To do so, we draw a set of outputs from the model, $\bd_1,\ldots,\bd_n \sim \PLLM(\cdot \vert \bp^*)$ and solve the maximum-likelihood problem where the objective is to find the prompt $\bp$ under which the example outputs are most likely to have been drawn.
\begin{align}\label{eq:mle-def-intro}
    \bp = \argmax_{\bp} \sum_i \log \PLLM(\bd_i \vert \bp).
\end{align}

This problem corresponds to optimizing an empirical approximation of the KL divergence between prompts $\bp$ and $\bp^*$, and is derived in Section~\ref{sec:reconstruct}.

In solving \eqref{eq:mle-def-intro}, the central obstacle is that prompts $\bp$ are discrete strings of tokens. Therefore, (\ref{eq:mle-def-intro}) is a discrete optimization problem and typical continuous optimization methods such as gradient descent do not apply. Instead, to perform this optimization, we build on methods developed in the adversarial attacks literature (see \citep{zou2023universal} and related work in Section~\ref{sec:related}).

\paragraph{Investigations on optimized prompts} We explore several interesting properties of these optimized prompts.
\begin{itemize}
    \item \textit{Evil twins}. In many cases, the optimized prompts that we find are similar in function to the original prompts (twins), but garbled and unintelligible to humans (evil). For this reason, we refer to them as \textit{evil twins}. See Figure~\ref{fig:prompt-optimized-examples} for some examples. 
    \item \textit{Transferability}. Remarkably, these ``evil twin'' prompts transfer between a variety of open-source and proprietary language models; see Section~\ref{sec:transferability}.
    \item \textit{Robustness}. We investigate the robustness of evil twin prompts to changes in their token-order and to replacements of their tokens. We find that whether evil twins are robust to randomly permuting their tokens depends on the LLM family. On the other hand, across LLM families, evil twins are more impacted by randomly replacing their tokens than ground truth prompts. This suggests that even the uncommon, non-English tokens in the optimized prompts play an important role in driving the model output; see Section~\ref{sec:prompt-investigations}.

    \item \textit{Improving prompt intelligibility}. We explore variants of the optimization problem \eqref{eq:mle-def-intro} that encourage the optimized prompts to be more interpretable (adding a fluency penalty and restricting the vocabulary to common English tokens). However, we find that these modifications do not improve the KL divergence of the optimized prompts to the ground truth; see Section~\ref{sec:optim-intel}.
\end{itemize}

We discuss other applications of the maximum-likelihood problem \eqref{eq:mle-def-intro} to prompt compression, privacy, and conditional generation in Section~\ref{sec:discussion}.

\section{Related work}\label{sec:related}

This paper fits into a quickly growing literature studying how language models parse prompts. Furthermore, the techniques used in this paper build off of a body of work on prompt optimization. We survey relevant work below.

\paragraph{How models parse prompts} There is rapidly mounting evidence that LLMs interpret natural-language prompts in counterintuitive ways. For instance, models struggle with prompts that are negated, such as prompts that ask to ``Give an \textit{incorrect} example'' instead of to ``Give a \textit{correct} example''~\citep{jang2023can}. Additionally, natural-language instructions in prompts in few-shot settings can often be replaced by irrelevant strings of text, with no drop in performance~\citep{webson-pavlick-2022-prompt}. Moreover, in few-shot settings the in-context examples' labels can be replaced by random labels with little drop in performance~\citep{min2022rethinking}. These experiments indicate that LLMs follow instructions in prompts differently than humans do, which agrees in spirit with our finding of evil twin prompts.

There is also existing evidence that LLMs are able to parse some non-natural language prompts. \citealp{daras2022discovering} finds that garbled text appearing in DALLE-2 images can be repurposed in prompts to the image generation model, and yields natural images. \citealp{milliere2022adversarial} suggests that this may be an artifact of the model's byte pair encoding, pointing out that the example prompt ``Apoploe vesrreaitais'', which generates bird images, is reminiscent of the real Latin bird families \textit{Apodidae} and \textit{Ploceidae}. Furthermore, adversarial example prompts that jailbreak models sometimes contain uninterpretable suffixes (e.g., \citep{cherepanova2024talking, zou2023universal,liu2023autodan}). Our results in this paper demonstrate that the phenomenon of language models parsing non-natural language prompts is more widespread than previously known, since many natural language prompts have non-natural language analogues. A full understanding of how models parse prompts will require contending with the existence of evil twin prompts.

\paragraph{Prompt optimization} 
The techniques in this work draw from the prompt optimization literature. This literature primarily includes optimization methods for \textit{hard prompts} (which are text strings, i.e., sequences of tokens), and \textit{soft prompts} (i.e., sequences of embedding vectors that are not constrained to correspond to a textual string). Hard prompts are more desirable because they are more easily inspected by humans, and can be inputted across different models. 

Foundational work for soft prompt optimization includes prefix tuning~\citep{li-liang-2021-prefix,lester2021prefix}, which trains a soft prompt with gradient descent. This soft prompt is then prepended to a hard prompt for improved conditional generation on a range of tasks. We include experiments on soft prompts in Appendix~\ref{app:soft-prompts}, but the focus of this paper is on hard prompts.

Hard prompt optimization operates in the model's discrete token space, meaning that the optimization is not directly differentiable.
Hard prompt optimization is most frequently described in the context of adversarial attacks or finding ``jailbreaks'' (prompts) that generate malicious output, or induce model misclassification. Several methods such as HotFlip~\citep{ebrahimi-etal-2018-hotflip}, AutoPrompt~\citep{shin-etal-2020-autoprompt}, Greedy Coordinate Gradient (GCG)~\citep{zou2023universal}, and AutoDAN~\citep{liu2023autodan} have been developed to optimize over hard prompts. These methods work by starting with an arbitrary prompt and iteratively modifying tokens towards the goal of obtaining the adversarial attack behavior. In our work, we apply GCG (plus extra warm starts, pruning, and fluency penalties) to our optimization framework, demonstrating that it can be used in settings beyond adversarial attacks.

The closest work to ours is PEZ~\citep{wen2023hard}, which proposes a method that takes input images and finds matching prompts in CLIP embedding space. This bears similarity to the maximum-likelihood problem in \eqref{eq:mle-def-intro}, but our setting differs significantly from PEZ in that our optimization problem does not rely on a multimodal model with a shared embedding space -- all that we require is the ability to compute the log-likelihood of a document given a prompt. In particular, our formulation of prompt optimization means that our method is applicable even when the documents outputted by the model do not have the same meaning as the prompt (i.e., the twin prompt does not have to be close to the documents in some embedding space). This is the setting in all conversational language models, where the model's responses are not paraphrases of the prompt.

\section{Preliminaries}
\label{sec:background}

\subsection{Autoregressive language models}
In our work, we focus on transformers \citep{vaswani2017attention} with a decoder-only architecture, as the majority of recent language models have adopted this architecture. 
We define a transformer language model $h$, with a vocabulary size of $V$ tokens, where each token maps to a $d$ dimensional embedding.
The input to the model is a length-$k$ sequence represented as a matrix $\bX \in \R^{k \times V}$ by stacking one-hot encodings $\bx_1,\ldots,\bx_k \in \R^V$ of tokens.

Given a sequence $\bX_{1:i} \in \R^{i \times V}$, the model outputs logits for the $(i+1)$ token probabilities $h(\bX_{1:i}) \in \R^V$. 

\subsection{Probability of a document}
Given the input sequence $\bX$, the model induces a probability distribution $\PLLM$ over the input:
\begin{align*}
    \PLLM(\bX) = \prod_{i=1}^k \bx_i^\top \smax(h(\bX_{1:(i-1)})),
\end{align*}
where $\bx_i$ is $i$th row of $\bX$, and for any vector $\bv \in \R^n$, the softmax is a vector in $\R^n$ given by $\smax(\bv)_i = \nicefrac{e^{v_i}}{\sum_{j=1}^n e^{v_j}}$.

Now, given an input sequence of a prompt concatenated with a document in the form
\begin{align*}
    \bX = [\bp, \bd] \in \R^{(k_p + k_d) \times V},
\end{align*}
where $\bp \in \R^{k_p \times V}$ and $\bd \in \R^{k_d \times V}$ are the prompt and document respectively, the conditional probability of the document given the prompt is
\begin{align}
\label{eq:prob-docs-definition}
\PLLM(\bd | \bp) = \prod_{i=k_p+1}^{k_p + k_d} \bx_i^{\top} \smax(h(\bX_{1:(i-1)})).
\end{align}

\section{Optimization problem}
\label{sec:reconstruct}

\subsection{KL divergence between prompts}
Given two prompts, $\bp, \bp^* \in \R^{k_p \times V}$, we use the KL divergence \eqref{eq:kl-defn-llm} to measure how the distributions over documents that the prompts induce differ. Since the KL divergence between distributions $f,g$ is defined as 
$$\KL(f||g):= \E_{x\sim f}[\log(f(x)) - \log({g(x)})],$$ our distance between prompts can be equivalently formulated as
\begin{align*}
    d_{KL}(\bp^* || \bp) =  \E_{\bd \sim \PLLM(\cdot | \bp^*)}[& \log(\PLLM(\bd | \bp^*)) \\ 
    & - \log(\PLLM(\bd | \bp))].
\end{align*}
Since we have access to the output log probabilities from the model, we can estimate the distance by drawing some number $n$ of documents $\bd_1,\ldots,\bd_n \sim\PLLM(\cdot|\bp^*)$  and computing 
\begin{multline} \label{eq:KL-prompts}
\hat{d}_{KL}^{(n)}(\bp^*||\bp) = \frac{1}{n} \sum_{i = 1}^n \log(\PLLM(\bd_i|\bp^*)) \\ 
    - \log(\PLLM(\bd_i|\bp)).
\end{multline}

As we increase $n$, the estimator $\hat{d}_{KL}^{(n)}$ concentrates around its expectation $d_{KL}$, and we obtain a good-quality approximation. We select the KL divergence as the statistical distance for prompt optimization because (i) it bounds the total variation distance by Pinsker's inequality \citep{pinskerinequality}, and, as we will now see, (ii) minimizing it naturally corresponds to maximum likelihood estimation, and (iii) it allows for efficient optimization.

\subsection{Optimization problem}
We seek a prompt $\bp$ that minimizes the empirical estimate of the KL divergence between $\bp^*$ and $\bp$ given in \eqref{eq:KL-prompts}. However, \eqref{eq:KL-prompts} involves additive terms that depend on $\bp^*$, which we cannot compute unless we know $\bp^*$. Fortunately, these terms do not depend on $\bp$, so in the optimization we can drop these terms and define the loss function
\begin{align*}
L(\bp;\bd_1,\ldots,\bd_n) = -\sum_{i=1}^n \log \PLLM(\bd_i | \bp),
\end{align*}
and the set of solutions remains unchanged
\begin{align}\label{eq:min-kl-rephrase}
\argmin_{\bp \in \cH} L(\bp; \bd_1,\ldots,\bd_n) = \argmin_{\bp \in \cH} \hat{d}_{KL}^{(n)}(\bp^* || \bp) \,.
\end{align}
Here $\cH$ is the set of hard prompts where each row of $\bp$ is a one-hot indicator vector for a token.

\textit{Remark}. As discussed in the introduction, the optimization problem that we solve corresponds to finding a maximum-likelihood estimator (MLE)
\begin{align*}
    \hbp^{MLE} &=  \arg\max_{\bp} \prod_{i=1}^n \PLLM(\bd_i | \bp) \nonumber\\
    &= \arg\max_{\bp}  \sum_{i=1}^n \log\PLLM(\bd_i | \bp) \\
    &= \argmin_{\bp} L(\bp; \bd_1,\ldots,\bd_n)\,,
\end{align*}
which is the prompt $\bp$ that maximizes the probability that the documents $\bd_1,\ldots,\bd_n$ are drawn.

\section{Comparison of optimization methods}\label{sec:methods-comparison}

We consider various methods to optimize \eqref{eq:min-kl-rephrase}.

\begin{itemize}
    \item \textit{Asking GPT-4}. Since this optimization is equivalent to the maximum-likelihood problem, we benchmark our methods against the ``optimization'' ability of commercial LLMs. Namely, we provide GPT-4 with our training corpus, containing the $n$ documents which are used for optimization, and ask it to provide an example prompt that could have generated the corpus; see Appendix~\ref{app:recon-examples} for more details and the GPT-4 prompt template.
    \item \textit{GCG with cold start}. We optimize \eqref{eq:min-kl-rephrase} with the Greedy Coordinate Gradient (GCG) algorithm \citep{zou2023universal}, which computes per-token gradients for each position in the prompt, and iteratively flips tokens in order to minimize the loss. The full GCG algorithm is reproduced in Appendix \ref{app:gcg}. In the \textit{cold start} version, we initialize a prompt $\bp^0 \in \R^{k_p \times V}$ to some arbitrary tokens from the vocabulary. 
    \item \textit{GCG with warm start}. We experiment with combining both of the above methods, by warm-starting the GCG algorithm using the suggested prompt from GPT-4.

    \item \textit{GCG with warm start, fluency penalty, and vocabulary pruning}. Since GCG (with both cold and warm starts) typically returns unintelligible prompts, we experiment with methods to get more interpretable prompts. These are presented and discussed in Section~\ref{sec:optim-intel}.
\end{itemize}

\begin{figure}[!b]
    \centering
\includegraphics[width=1.0\linewidth]{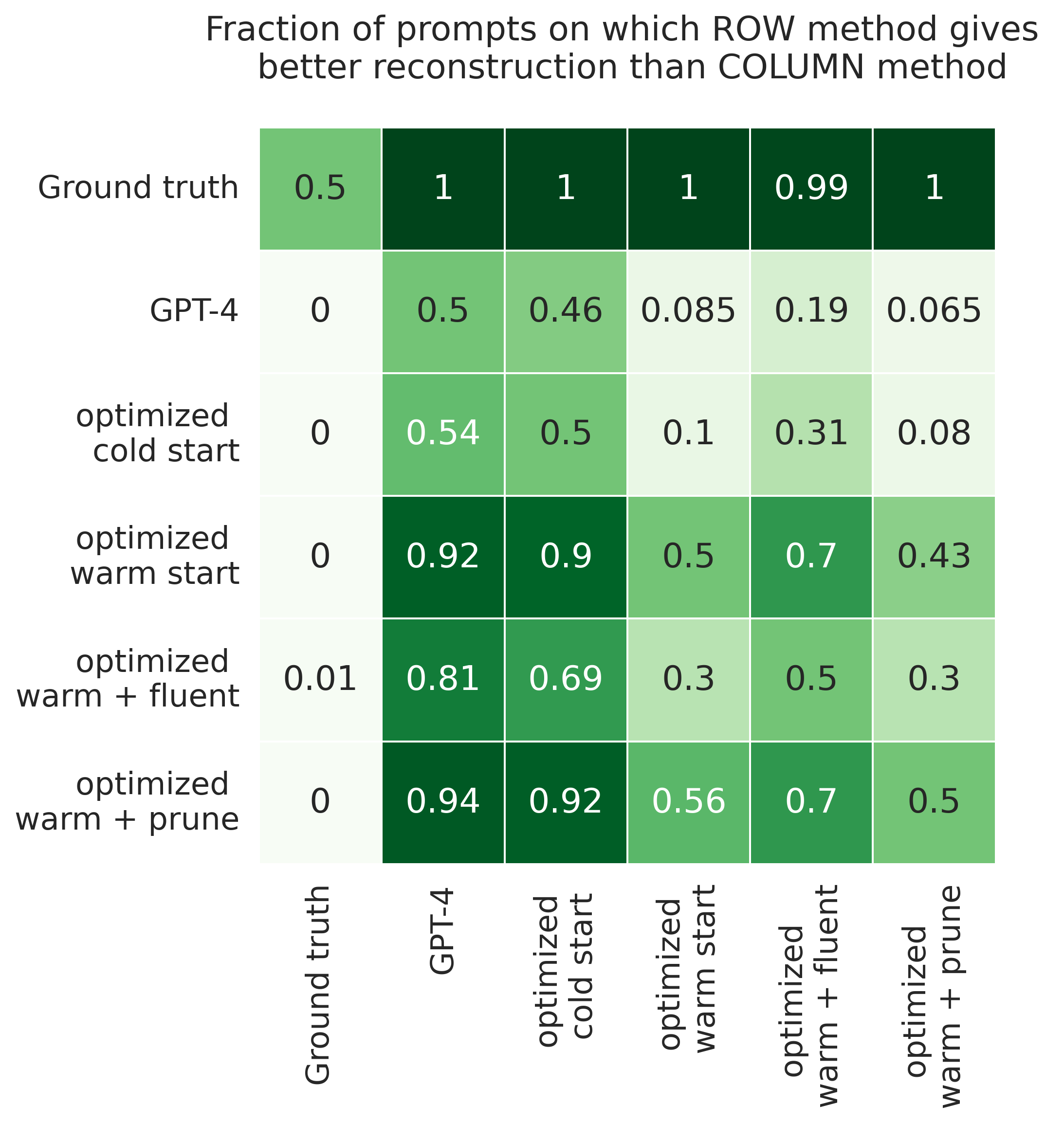}
    \caption{Win rate between various methods across optimizations of 100 ground truth prompts with 100 documents each. Given two prompts to compare, we compute the KL divergence for both prompts with respect to the ground truth, and the method with lower KL wins. Darker shades indicate ROW method is better than COLUMN method. Full optimization results are shown in Appendix~\ref{app:recon-examples}. In the case of ties, the win is shared by both methods. The most effective method is GCG with warm starts.}
    \label{fig:comp-techniques}
    \vspace{-.3cm}
\end{figure}

We compare these methods on 100 randomly sampled prompts from the Alpaca instruction tuning dataset \citep{alpaca}, where Vicuna-7b-v1.5 is the instruction-tuned model. Additional experiments on various model families and datasets are presented in Appendix~\ref{app:addn-model-exp}. For each method and prompt, we compute the KL divergence of the optimized prompt with respect to the original prompt. We compare pairs of methods based on which one finds the closer prompt to the ground truth; see Figure~\ref{fig:comp-techniques}. GPT-4 suggestions perform roughly on par with those from cold-start GCG. On the other hand, GCG with a warm start provides a strong improvement over both cold-start GCG and the GPT-4 prompt suggestions. Enforcing interpretability by adding a fluency penalty or pruning the vocabulary does not improve the optimized prompt (see Section~\ref{sec:optim-intel}). All results are reported in Figure~\ref{fig:full-kl-results}.

\section{Evil twin prompts transfer between models}\label{sec:transferability}

We test whether prompts optimized on one model work on other models from different families and of different sizes.

\subsection{Transferability to open source and proprietary models}\label{sec:transfer-closed}

Although the optimized ``evil twin'' prompts are generally unintelligible to humans, we surprisingly find that they transfer to a number of open source and closed industrial LLMs. We use 100 optimized (from a GPT-4 warm start) prompts from Vicuna and run them through a variety of open source and closed models. We use GPT-4 as a judge to determine if the induced responses from the optimized prompt are faithful to the original prompt on a scale of 1 to 3. 

\begin{table}[h!]
\centering
\small
\setlength{\tabcolsep}{4pt} 
\begin{tabular}{c|ccc}
\toprule
Model & Score = 1 & Score = 2 & Score = 3 (best)\\
\midrule
Gemini Pro & 17 & 8 & \textbf{75} \\
GPT-3.5-turbo  & 31 & 6 & \textbf{63} \\
GPT-4  & 31 & 7 & \textbf{62} \\
Claude 3 Haiku & \textbf{59} & 5 & 36 \\
Claude 3 Sonnet  & 38 & 8 & \textbf{54} \\
mistral-medium  & 16 & 30 & \textbf{54} \\
mistral-small  & 21 & 12 & \textbf{67} \\
mistral-tiny  & 24 & 22 & \textbf{53} \\
OpenHermes-2.5 & 5 & 24 & \textbf{71} \\
OpenHermes-13B  & 28 & 19 & \textbf{53} \\
Llama2-7b-chat & 7 & 28 & \textbf{64} \\
Llama2-13b-chat & 8 & 27 & \textbf{64} \\
Vicuna-7B & 7 & 22 & \textbf{71} \\
Vicuna-13B & 8 & 27 & \textbf{64} \\
\bottomrule
\end{tabular}
\caption{Transferability results to open source and proprietary models. Using 100 optimized prompts from Vicuna, we directly input these prompts to various open source and closed models. The ratings are given by GPT-4, based on the scale described in the prompt in Section~\ref{sec:transfer-closed}.}
\label{tab:api-transfer}
\end{table}

Specifically, the prompt that we use for GPT-4 is:

{\small\texttt{Please judge if the following response answers the prompt. Use a scale of 3 rating, where: 1 means that the response does not answer the prompt at all, and is completely wrong; 2 means that the response gets the general idea of the prompt and answers it to some extent; and 3 means that the response faithfully answers the prompt.}}

Our results are shown in Table~\ref{tab:api-transfer}. We find that for all models (except Claude 3 Haiku), over 50\% of optimized prompts transfer with the highest rating.
Figure~\ref{fig:gemini-transfer} shows a visual example of transferability to the commercial Google Gemini Pro LLM.

\subsection{Transferability between model sizes}\label{sec:transfer-sizes}

\begin{figure}[tbp]
    \centering
    \includegraphics[width=1.0\linewidth]{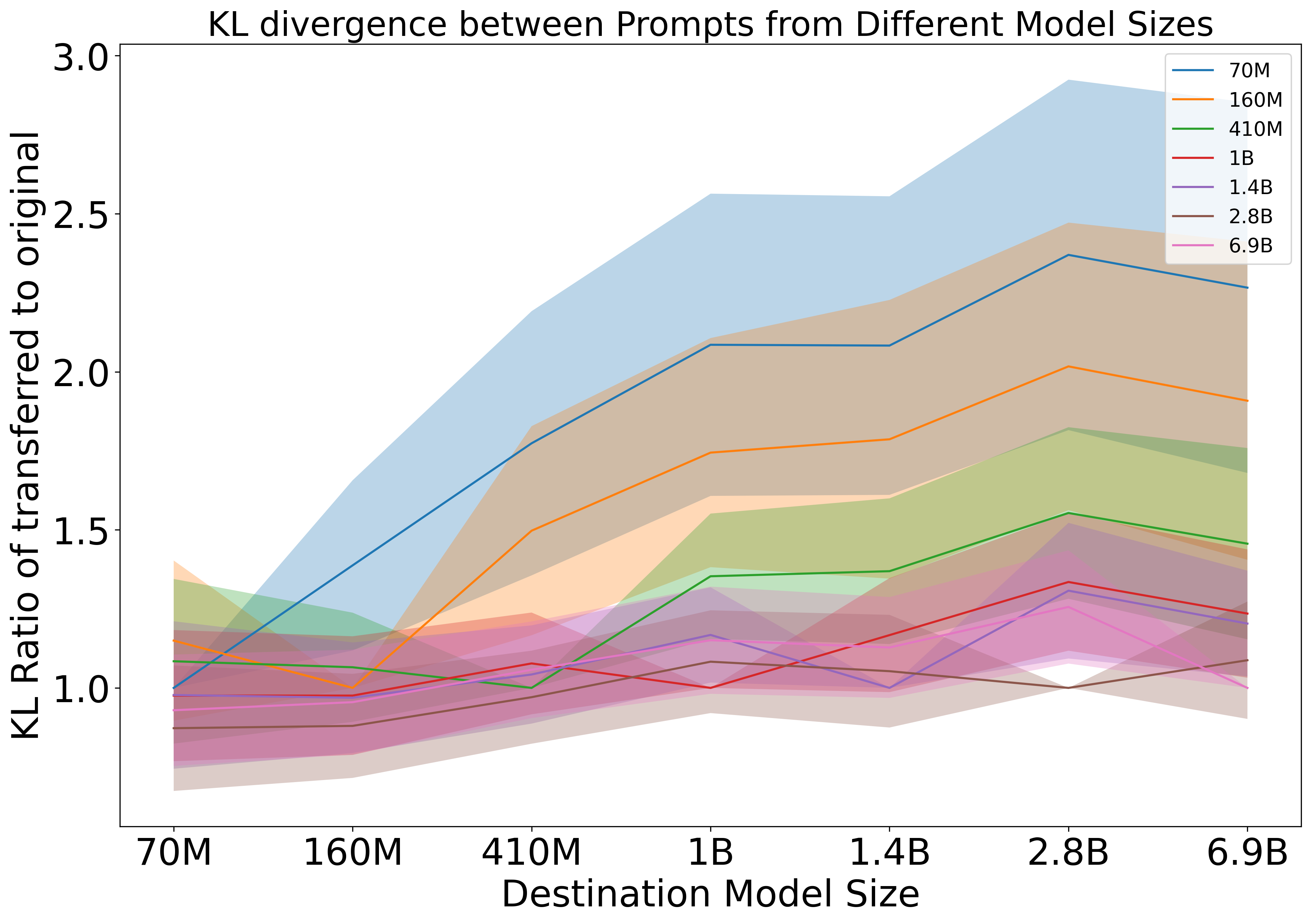}
    \caption{Transferability between model sizes. For each model size in the Pythia suite (excluding 12B), and each of 100 prompt sentences from the HellaSwag dataset \citep{zellers-etal-2019-hellaswag}, we run GCG with cold start to generate an optimized prompt based on 100 documents from the original prompt. For each optimized prompt at each model size, we compute the KL divergence for the optimized prompt at all other model sizes. The measured ratio is 
    $\frac{d_{KL,\mathrm{dest}}(\bp^* \parallel \bp_{\mathrm{source}})}{{d_{KL,\mathrm{source}} (\bp^* \parallel \bp_{\mathrm{source}})}}$ 
    averaged over all 100 prompts, where $\bp_{\mathrm{source}}$ represents the optimized prompt from the source model, $d_{KL,\mathrm{source}}$ represents the KL divergence as measured on the source model, and $d_{KL,\mathrm{dest}}$ represents the KL divergence as measured on the destination model.
    Full results are shown in Table~\ref{tab:transfer-results-pythia}.
    }
    \label{fig:transfer-ratio}
\end{figure}

Next, we study the transferability of optimized prompts between different models within a model family while varying the size. The Pythia \citep{biderman2023pythia} suite includes models ranging from 70M to 12B parameters. 
Each model is identical apart from the number of parameters, which makes it ideal for investigating how the distance between prompts changes with model size. Additionally, each model is trained with the same data seen in the same order. Our results are shown in Figure \ref{fig:transfer-ratio}. We find that prompts optimized on smaller models have worse transferability to larger ones. However, prompts optimized on larger models transfer very well to smaller ones.

\section{Robustness of optimized prompts}\label{sec:prompt-investigations}
\subsection{Token order sensitivity}\label{sec:order-sensitivity}

Natural language is sensitive to token order, in that the meaning of a sequence can be affected by re-arrangement of its constituent tokens. \citealp{ishibashi-etal-2023-evaluating} finds that prompts learned by AutoPrompt are more sensitive to token rearrangement than prompts written manually, as measured by performance on natural language inference tasks. We examine whether this is also true of our optimized prompts, invoking a KL-based assessment:

\begin{definition} Given prompts $\ba$ and $\bb$, define $\tba,\tbb$ to be random prompts formed by uniformly shuffling their tokens. We say that prompt $\ba$ is more \textbf{token-order-sensitive} than $\bb$ if
\begin{align*}
\P_{\tba,\tbb}(d_{KL}(\ba || \tba) > d_{KL}(\bb || \tbb)) > 0.5\,.
\end{align*}

\end{definition}

We wish to compare the token-order-sensitivity of optimized prompts to that of the natural-language ground truth prompts. We evaluate this using Algorithm \ref{alg:dominace}, which calculates a token-order-sensitivity ``win rate'' $w$ between $\bp$ and $\bp^*$, comparing how much the prompts change under random token reordering.

 \begin{algorithm}
\caption{Token-Order-Sensitivity Test}\label{alg:dominace}
\begin{algorithmic}[1]
\Require Number of trials $m$. Number of documents to generate $g$. Number of prompt pairs $n$.
\Ensure Test statistic $U$.

    \State $U \gets 0$
    \For{each $(\bp^*,\bp)$}
        \State $w \gets 0$
        
        \For{$i = 1$ to $m$}
            \If{$\hat{d}_{KL}^{(g)}(\bp||\tbp) < \hat{d}_{KL}^{(g)}(\bp^*||\tbp^*)$}
                \State $w \gets w + 1/m$
            \EndIf
        \EndFor

        \State $U \gets U + \frac{1}{n} (\mathbf{1}_{\{w > 0.5\}} + \frac{1}{2} \cdot \mathbf{1}_{\{w = 0.5\}})$
    \EndFor
    \Return $U$
\end{algorithmic}
\end{algorithm}

\begin{table}[h!]
\centering
\small
\setlength{\tabcolsep}{4pt} 
\begin{tabular}{c|cc}
\toprule
Model & $U$ & $w$ \\
\midrule
pythia-70m & $1.00$ ($0.95$, $1.00$) & $0.93$ ($0.85$, $0.96$) \\
pythia-160m & $1.00$ ($0.95$, $1.00$) & $0.97$ ($0.92$, $0.99$) \\
pythia-410m & $1.00$ ($0.96$, $1.00$) & $0.99$ ($0.93$, $0.99$) \\
pythia-1b & $1.00$ ($0.96$, $1.00$) & $0.99$ ($0.95$, $1.00$) \\
pythia-1.4b & $1.00$ ($0.95$, $1.00$) & $0.99$ ($0.93$, $0.99$) \\
pythia-2.8b & $1.00$ ($0.96$, $1.00$) & $0.99$ ($0.93$, $0.99$) \\
pythia-6.9b & $1.00$ ($0.96$, $1.00$) & $0.99$ ($0.95$, $1.00$) \\
vicuna-7b (cold) & $0.52$ ($0.42$, $0.62$) & $0.54$ ($0.43$, $0.63$) \\
vicuna-7b (warm) & $0.39$ ($0.29$, $0.48$) & $0.41$ ($0.31$, $0.50$) \\
gemma-2b-it (cold) & $0.63$ ($0.52$, $0.71$) & $0.59$ ($0.48$, $0.67$) \\
gemma-2b-it (warm) & $0.84$ ($0.74$, $0.89$) & $0.67$ ($0.57$, $0.75$) \\
mistral-7b-ins (warm) & $0.25$ ($0.17$, $0.33$) & $0.32$ ($0.24$, $0.42$) \\
phi-2 (warm) & $0.97$ ($0.92$, $0.99$) & $0.94$ ($0.86$, $0.97$) \\
\bottomrule
\end{tabular}
\caption{Token-order-sensitivity results. Given 100 prompt pairs $(\bp^*,\bp)$, we apply Algorithm \ref{alg:dominace} to assess token-order-sensitivity. Warm indicates that the optimized prompt was warm-started, while cold indicates that the optimized prompt was arbitrarily started. All runs of GCG on Pythia models were cold-started. The value of $U$ indicates the fraction of ground-truth prompts $\bp^*$ that are more token order sensitive than the corresponding optimized prompts $\bp$. We also report the average of win rates $w$ across prompt pairs and shufflings. Intervals for $U$ and $w$ reflect $95\%$ Clopper-Pearson intervals for binomial proportions \citep{clopper1934use}.}
\label{tab:shuffle-results}
\end{table}

\begin{figure*}[h!]
    \centering
    \includegraphics[width=1.0\textwidth]{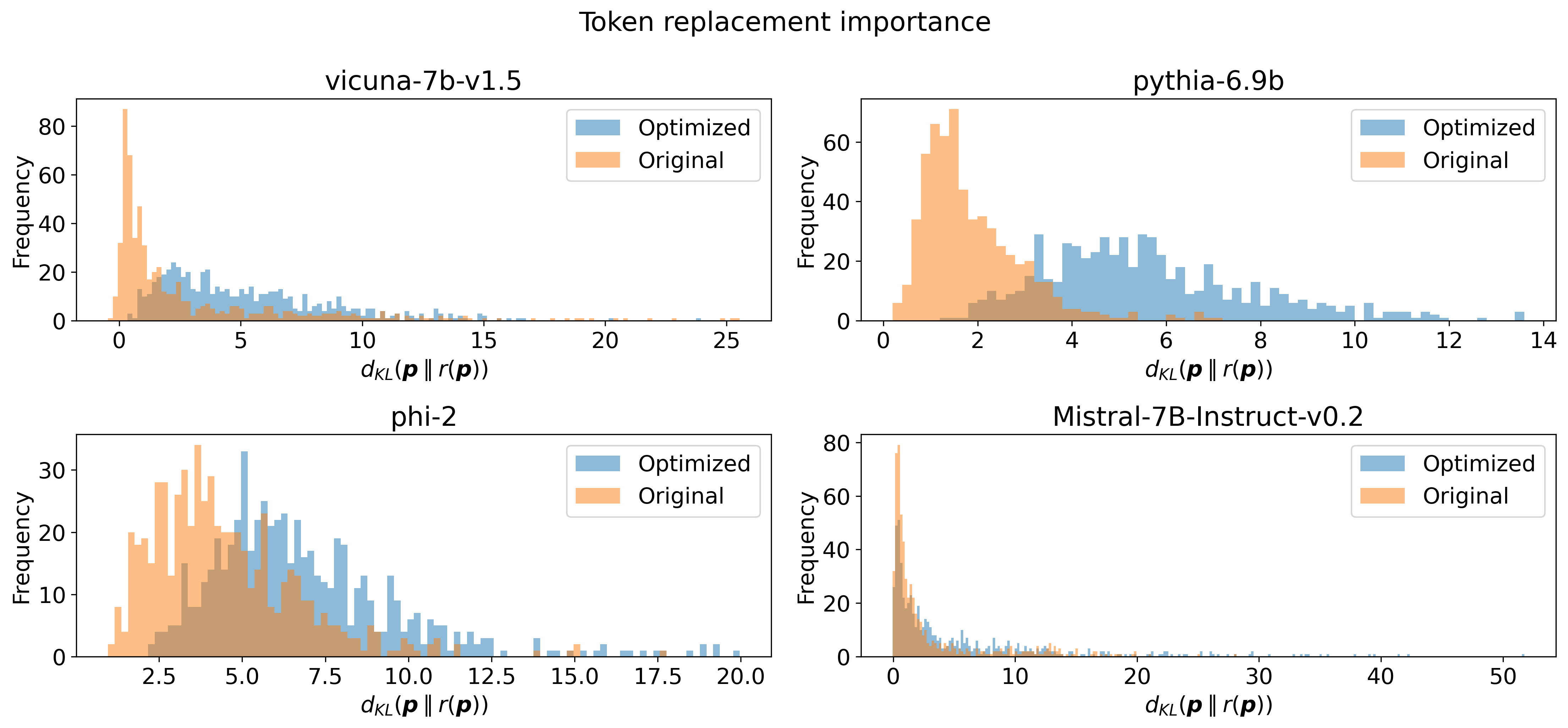}
    \caption{Individual token importance in optimized and original prompts for various models. For each of the 100 prompts from the Alpaca \citep{alpaca} and OpenHermes-2.5 datasets, and for each of the first 6 positions $i \in \{1,\ldots,6\}$ of the prompt, we compute the KL divergence $d_{KL}(\bp \parallel r_i(\bp))$ when we replace position $i$ with the [UNK] token. Each histogram is over all positions and prompts (either the original prompts or optimized prompts) for a given model. The optimized prompts appear to be generally more sensitive.}
    \label{fig:token-importance-vicuna-pythia}
\end{figure*}

We find that token order sensitivity appears to be dependent on the model family; see Table~\ref{tab:shuffle-results}. For Pythia, Phi-2 and Gemma, the optimized prompts are significantly less order sensitive than the ground truth prompts. For Mistral, the optimized prompts are somewhat more order sensitive. And for Vicuna, there is no significant difference between optimized and ground truth prompts.

\subsection{Token replacement sensitivity}\label{sec:token-replacement}

Based on visual inspection of the evil twin prompts in Figures~\ref{fig:prompt-optimized-examples} and \ref{fig:full-kl-results}, one can hypothesize that these consist of some tokens that are highly-related to the ground truth prompts and that drive the model's output, as well as some tokens that appear unrelated and can be safely ignored or replaced.

We test this hypothesis quantitatively, checking whether there are a few tokens in the optimized prompts that have an outsized effect on the prompt's functionality. We compute $d_{KL}(\bp || r_i(\bp))$ for each optimized prompt $\bp$, where $r_i$ is a function that replaces the $i^{th}$ token of a sequence with [UNK]. We do the same for the ground truth prompts $\bp^*$. Figure \ref{fig:token-importance-vicuna-pythia} plots histograms of these KL divergences over all prompts and token positions $i$.

Surprisingly, this experiment contradicts the hypothesis. Figure~\ref{fig:token-importance-vicuna-pythia} shows that the effect of replacing a token in the optimized prompts with the ``unknown'' token, [UNK], is generally \textit{greater} than the effect of replacing a token with [UNK] in the ground truth prompts. Thus, optimized prompts are more dependent on all of their tokens being present in a way that natural prompts are not, even though many of these tokens may appear garbled and uninterpretable. This effect is especially significant in the Pythia, Vicuna, and Phi-2 models, since very few tokens in the optimized prompts yield zero KL divergence change when they are replaced by [UNK].

\section{Optimizing for more intelligible prompts}\label{sec:optim-intel}

The prompts generated by our optimization are often unintelligible, and it may be desirable to recover a prompt that is more interpretable by humans. In this section, we explore two adjustments to our optimization procedure that aim to improve intelligibility: (1) fluency penalty, and (2) limiting the optimized prompt's vocabulary to common English tokens. We find that these variants do not improve the KL divergence of the optimized prompt to the original.

\subsection{Fluency penalty}
Inspired by prior work \citep{guo-etal-2021-gradient,mehrabi-etal-2022-robust,shi2022toward,wen2023hard} on adding additional terms such as perplexity, BERTscore \citep{zhang2019bertscore} and a fluency penalty to the loss in order to improve downstream performance, we follow \citep{shi2022toward} and add a term to the hard prompt loss function in order to penalize the log-likelihood of the prompt (fluency penalty). 
Our hard prompt loss function then becomes
\begin{align*}
    L(\bp;\bd_1,\ldots,\bd_n) = -\frac{1}{n} \sum_{i=1}^n \log \PLLM(\bd_i | \bp)  \\
    +\gamma \log \PLLM(\bp)
\end{align*}
where $\gamma \geq 0$ is a parameter controlling the importance of recovering a natural prompt. 
Larger $\gamma$ biases the optimization towards more natural prompts that may not necessarily fit the documents as well. We find that adding the fluency penalty decreases the similarity between the optimized and ground truth prompt; see Figure~\ref{fig:comp-techniques}. However, the prompts generated with a fluency penalty contain fewer strange tokens, and have higher fluency; see Figure~\ref{fig:full-kl-results} for the full results. An analysis of tuning the fluency hyperparameter $\gamma$ is provided in Appendix~\ref{app:fluency-analysis}.

\subsection{Vocabulary pruning}
We explore limiting the tokens chosen for GCG in order to improve reconstruction and fluency. Since all of our testing is carried out on English prompts and documents, we focus on English sub-words in the tokenizer only. In order to achieve this, we run the Llama tokenizer on an English corpus obtained from spaCy \citep{spacy2}, and mask out all tokens that do not appear in the corpus. The Llama tokenizer contains 32,000 tokens, and our pruning procedure results in about 15,000 tokens being removed.

We find that overall vocabulary pruning does not improve performance for reconstruction in a statistically significant manner across the 100 ground-truth prompts, although it does make the optimized prompts have fewer special characters; see Figure~\ref{fig:comp-techniques} and the optimization results in Figure~\ref{fig:full-kl-results}.

\section{Discussion and future work}\label{sec:discussion}

Our work takes a new perspective 
on prompt optimization by inquiring whether we can optimize prompts to be functionally equivalent to a certain ground-truth prompt. Functional similarity is quantified via the KL divergence between the ground truth prompt distribution and the optimized prompt's distribution. This yields a maximum-likelihood problem \eqref{eq:mle-def-intro}, whose solution uncovers ``evil twin'' prompts. Beyond our explorations of the transferability between models and robustness to perturbations of evil twin prompts, there are several open directions for future work. These directions include applications of the maximum-likelihood problem \eqref{eq:mle-def-intro} that are of independent interest.
\begin{itemize}

    \item \textit{Prompt compression}. By adding a length penalty to the optimized prompt in \eqref{eq:mle-def-intro}, our framework can be used to generate shorter prompts that mimic an original, longer prompt, which can then be used for pay-by-token API services in order to reduce inference time, context length usage, and total costs.
    \item \textit{Conditional generation}. The maximum-likelihood problem \eqref{eq:mle-def-intro} can be extended to prompts that allow for conditional generation. An example of where this may be useful is in style/content transfer: given a set of user emails in the form (topic, email), a user could optimize a prompt such that the concatenated input string [prompt; topic] would be likely to generate the corresponding emails, and could write new e-mails on new topics in the user's style as defined by the user's corpus of previous e-mails.
    \item \textit{Corpus compression}. One could apply our framework \eqref{eq:mle-def-intro} to help compress corpora of documents. Given documents $\bd_1,\ldots,\bd_n$ drawn from a distribution, one would find an optimized prompt that would configure the model to be better at predicting documents from that distribution. This could yield improved performance if the model were used as a compression algorithm via arithmetic encoding as in \citep{deletang2023language}.
\end{itemize}

\section*{Limitations}

The evil twins that we find are discovered using the GCG algorithm \citep{zou2023universal} plus additional warm-starting, token pruning, and fluency penalties. However, GCG may not result in a stable optimization in all cases. This can be seen in Appendix~\ref{app:recon-examples}, where for some examples the optimization fails to find prompts with low KL divergence to the original prompt. Thus, in the future it makes sense to explore alternative optimization algorithms, such as algorithms that may edit not just one token at a time, but may also make multi-token insertions and deletions, as well as vary the number of tokens during the optimization. Also, additional future work is required to adapt our framework for the applications of independent interest, because GCG may take many iterations to converge, which may introduce a significant runtime overhead.

Our approach for finding evil twins relies on having full access to the model's gradients, which is not the case for many closed-source models such as GPT-4. Nevertheless, the transferability of evil twins between models allows us to find them on open-source models and apply them to closed-source models.

\section*{Potential risks}

It is possible for a malicious user to use our framework to construct a prompt that generates a corpus of toxic or harmful documents, while not appearing malicious at surface level. However, there are many ways to mitigate the risks, such as perplexity filters and prompt paraphrasing \citep{jain2023baseline}.

\section*{Acknowledgements}

This research was developed in part with funding from NSF under grant 2127207. EB was funded by NSF grant 1745302.

\bibliography{anthology,custom}

\appendix

\algnewcommand{\LeftComment}[1]{\Statex #1}

\begin{algorithm}
\caption{Greedy Coordinate Gradient (GCG)}\label{alg:gcg}
\begin{algorithmic}
\Require Initial prompt $\bX_{1:n}$, loss $\cL$
\Ensure Optimized prompt
\For{$T$ epochs} 
\For{$i \in \{1,\ldots,n\}$} \LeftComment{\quad ~~~~~ {\color{brown}// Compute promising token substitutions}}
\State $\cX_i := \mathsf{TopK}(-\nabla_{e_{\bx_i}}\cL(x_{1:n}))$ 
\EndFor
\For{$j \in \cX_i$}
\State $\overline{\bX}^{(j)}_{1:n}:=\bx_{1:n}$
\State $\overline{\bx}^{(j)}_{i}:=\Unif(\cX_j)$
\EndFor   
\LeftComment{\quad ~~{\color{brown}// Compute best replacement}}
\State $j^\ast = \argmin_j \cL( \overline{\bX}^{(j)}_{1:n})$
\State $\bX_{1:n}:= \overline{\bX}^{(j^\ast)}_{1:n}$
\EndFor
\end{algorithmic}
\end{algorithm}

\section{Greedy Coordinate Gradient algorithm}\label{app:gcg}
Our paper builds on the Greedy Coordinate Gradient (GCG) algorithm from \citep{zou2023universal} for prompt optimization given in Algorithm \ref{alg:gcg}, by incorporating warm starts and experimenting with vocabulary pruning. GCG falls in a line of discrete optimization algorithms that iteratively construct prompts using token flips, combined with various heuristics for which tokens to flip and in what order. 

Early work, such as HotFlip \citep{ebrahimi-etal-2018-hotflip}, picks a token and approximates the top-1 token in the vocabulary which decreases the loss most when flipped to. This is able to induce incorrect classification for sentiment analysis.

Building on this, AutoPrompt appends a small number of randomly initialized "trigger" tokens to the original prompt. The tokens in this "trigger" are subsequently masked and optimized via masked language modeling, where the objective is to minimize the loss of the input sequence by by selecting some top-$k$ tokens with highest gradient for each trigger \citep{shin-etal-2020-autoprompt}.

GCG utilizes a similar approach to AutoPrompt; given a suffix of tokens to the task prompt, they optimize this suffix by a computing the top-$k$ tokens with largest negative gradients for every position in the suffix, then uniformly sample a single token as a candidate replacement for each position in the suffix. Finally, for each candidate suffix, they compute the loss by running a forward pass, and select the candidate suffix with lowest loss as the final new suffix. Using their optimized suffixes, they are able to generate prompts which induce malicious output from open source LLMs such as Llama, as well as large commercial models such as ChatGPT and GPT-4. The full algorithm details for GCG are shown in Algorithm \ref{alg:gcg}.

\begin{figure} %
    \centering
    \includegraphics[width=0.8\linewidth]{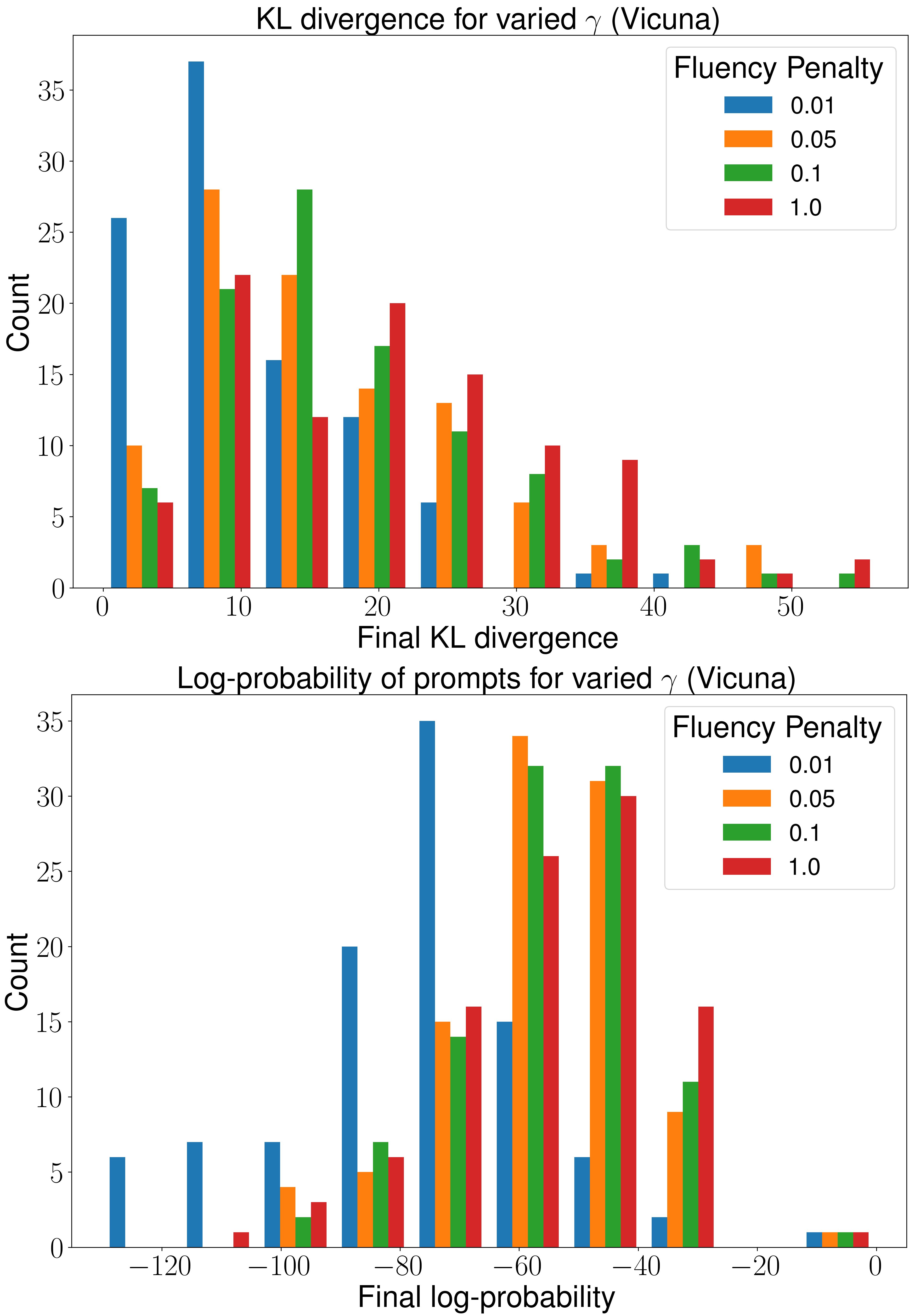}
    \caption{Hard prompt optimization results for various fluency penalties $\gamma$ with the Vicuna-7b model. We use a 100 prompt subset from Alpaca, and Vicuna-7b from a GPT-4 warm start. The optimization proceeds for 50 epochs, and we take the final values of the KL divergence to the ground truth, and the log-probability of the optimized prompt. %
    }
    \label{fig:fluency-search-vicuna}
\end{figure}

\begin{figure} %
    \centering
    \includegraphics[width=0.8\linewidth]{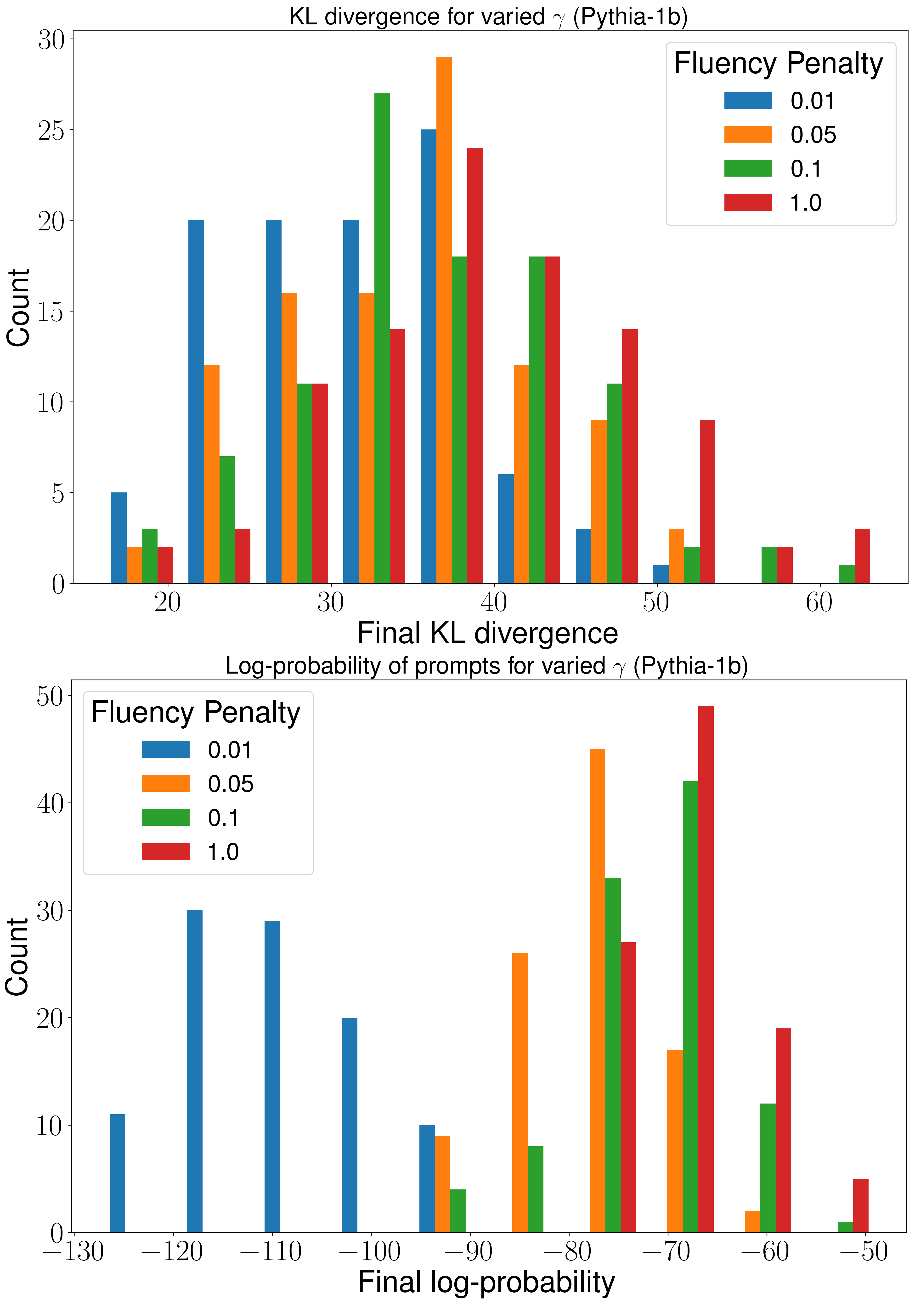}
    \caption{Hard prompt optimization results for various fluency parameters $\gamma$ with the Pythia-1b model. We use a 100 prompt subset from HellaSwag, and Pythia-1b with a cold start. The optimization proceeds for 50 epochs, and we take the final values of the KL divergence to the ground truth, and the log-probability of the optimized prompt.}
    \label{fig:fluency-search-pythia}
\end{figure}

\section{Fluency hyperparameter analysis}\label{app:fluency-analysis}

We explore the effects of varying the strength of the fluency penalty by selecting $\gamma \in \{0.01, 0.05, 0.1, 1.0\}$ and running hard prompt optimization for 50 epochs on Vicuna-7b with a GPT-4 warm start; see Figure~\ref{fig:fluency-search-vicuna}. We also run hard prompt optimization on Pythia-1b for 50 epochs from a cold start; see Figure~\ref{fig:fluency-search-pythia}.

These figures show a perhaps surprising trade-off between the readability of the prompt (as measured by the final log probability), and how well it reconstructs the original prompt. For our optimizations in Figure~\ref{fig:comp-techniques}, we select $\gamma = 0.05$, and this value does degrade the optimization performance in terms of KL divergence to the ground truth.

\begin{figure}[tbp]
    \centering
    \includegraphics[width=1.0\linewidth]{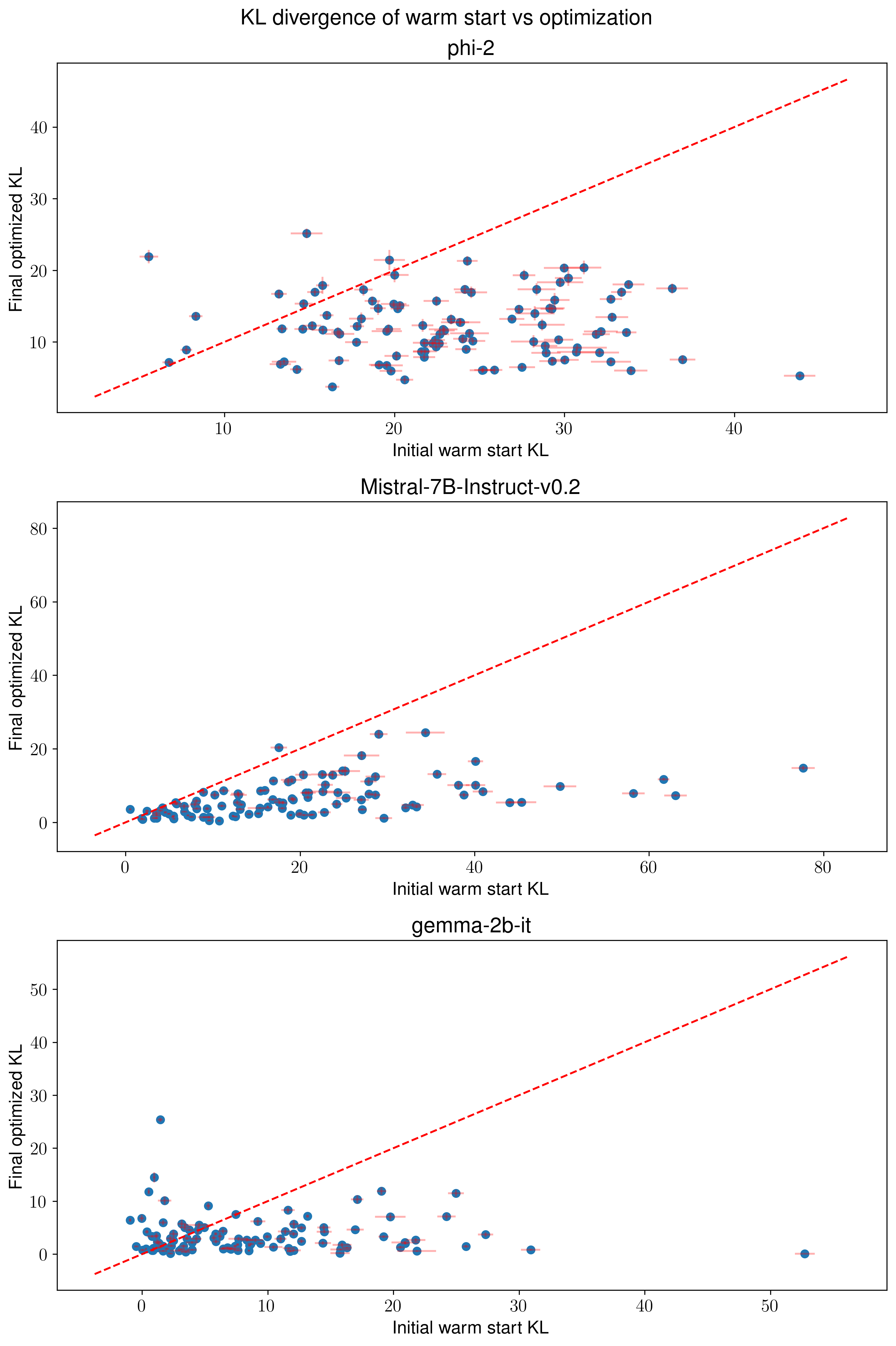}
    \caption{Hard prompt optimization with Phi-2, Mistral-7B-Instruct, and Gemma-2B. 100 prompts are randomly sampled from a subset of the OpenHermes-2.5 dataset which involves coding tasks, and we run hard prompt optimization for 100 epochs, beginning with a warm-start from GPT-4. Each point is one prompt. Horizontal error bars capture uncertainty for the initial warm start KL, while vertical error bars capture uncertainty in the final optimized KL.}
    \label{fig:addn-models-kl}
\end{figure}

\section{Additional experiments with varied model families and datasets}\label{app:addn-model-exp}

We run additional experiments on Microsoft's \href{https://huggingface.co/microsoft/phi-2}{Phi-2} (2.7 billion parameters), Mistral's \href{https://huggingface.co/mistralai/Mistral-7B-Instruct-v0.2}{Mistral-7B-Instruct-v0.2} (7 billion parameters), and Google's Gemma (2 billion parameters)~\citep{gemmateam2024gemma}. We use the popular prompt dataset \href{https://huggingface.co/datasets/teknium/OpenHermes-2.5}{OpenHermes-2.5}, which contains a diverse variety of prompts for various tasks such as coding, Q\&A, and many others. We filter for a subset of prompts that are related to writing code. 

For all models, we run hard prompt optimization for 100 epochs, starting from a GPT-4 warm start. We find that we achieve similar results as we did with other model families; see Figure~\ref{fig:addn-models-kl}.

\section{Soft prompt results}\label{app:soft-prompts}

Each token in the vocabulary $V$ maps to a $d$ dimensional embedding. We denote the embedding layer by $\bW_E \in \R^{V \times d}$, meaning that the model is in the form $h(\bX) = g(\bX \bW_E)$, where $g$ is the rest of the transformer model except the embedding layer.

\begin{figure} %
    \centering
    \begin{tabular}{@{}c@{}}
    \includegraphics[width=0.8\linewidth]{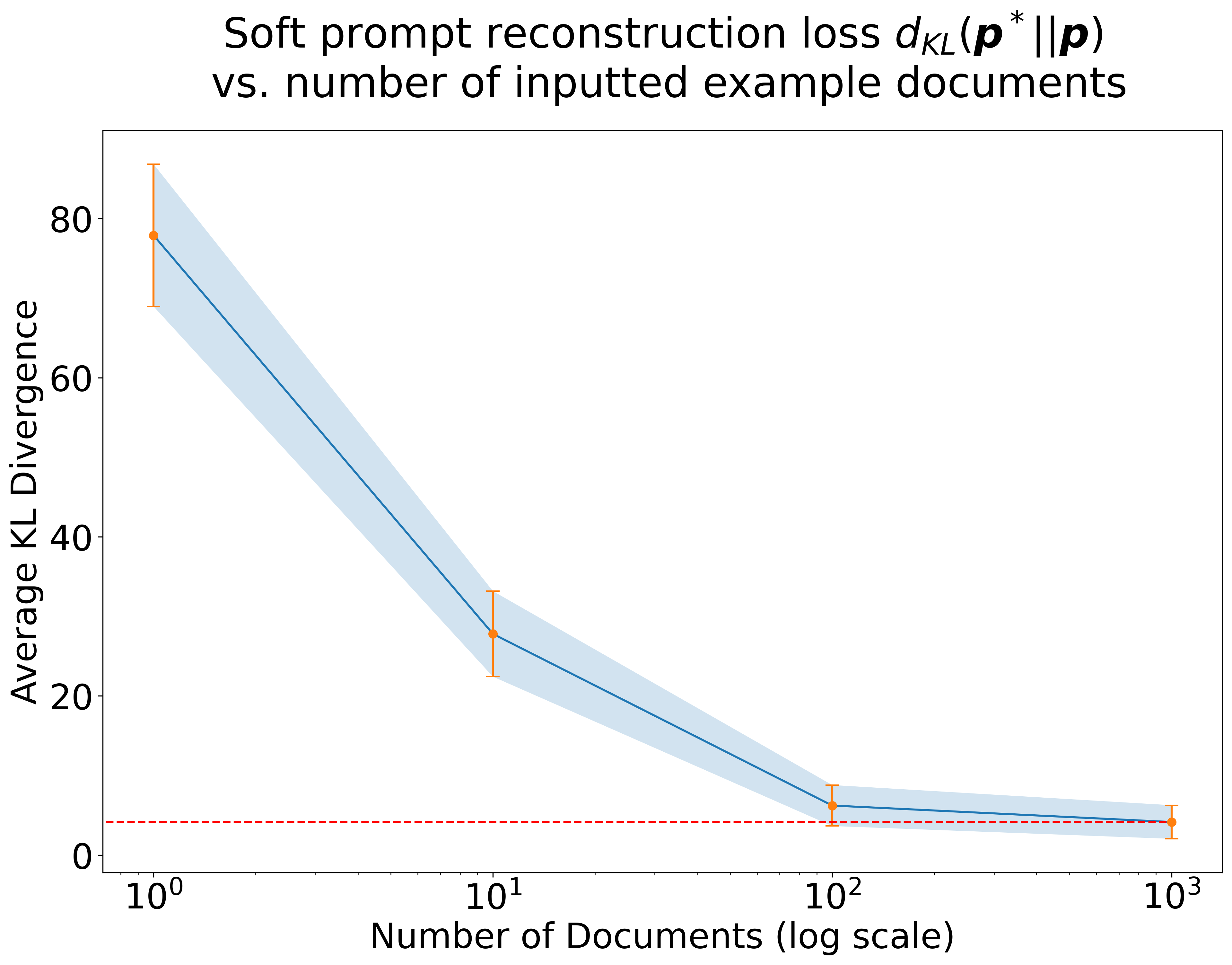}
    \end{tabular}
    \caption{Using Pythia 1.4b and a single prompt $\bp^*$, we generate sets of documents of varying sizes. For each set, we run soft prompt reconstruction, and report the KL divergence with $\bp^*$ and select the best value out of 200 epochs. Error bars capture the uncertainty over 3 trials plus uncertainty in the KL approximation on the held-out set of 100 documents.}
    \label{fig:soft-prompt-reconstruction}
\end{figure}

Recall that \textit{soft prompts} are sequences of vectors that lie in $\R^d$ where $d$ is the dimensionality of the embedding space, rather than sequences of tokens. Specifically, we can represent the soft prompt as a matrix $\bZ \in \R^{k_p \times d}$, which is fed into the LLM instead of the prompt's embeddings, and similarly to \eqref{eq:prob-docs-definition} induces a distribution over documents $\bd \in \R^{k_d \times V}$. In a slight abuse of notation:
\begin{align*}
    \PLLM(\bd | \bZ) = \prod_{i=1}^{k_d} \bd_i^{\top} \smax(g(\bX_{1:(k_p+i-1)})), \\
    \quad \bX = [\bZ, \bd \bW_E] \in \R^{(k_p + k_d) \times d}.
\end{align*}
Thus, we can use the MLE formulation as defined in \eqref{eq:min-kl-rephrase} with loss function
\begin{align*}
 L(\bZ;\bd_1,\ldots,\bd_n) = -\frac{1}{n} \sum_{i = 1}^n \log \PLLM(\bd_i | \bZ).
\end{align*}
The vectors in soft prompts do not have to correspond to embeddings of tokens, which makes the optimization problem \eqref{eq:min-kl-rephrase} continuous. This means that we can optimize the prompt $\bp$ by running gradient descent (GD), where we initialize $\bZ^0$ with random embedding vectors on each row, and $\eta > 0$ is a step size
\begin{align*} \tag{GD on prompt embeddings}
    \bZ^{t + 1} = \bZ^t - \eta \nabla_{\bZ} L(\bZ;\bd_1,\ldots,\bd_n)\,.
\end{align*}
In Figure~\ref{fig:soft-prompt-reconstruction}, we plot the results of soft-prompt reconstruction with varying numbers of documents. As the number of documents increases, the recovered soft prompt converges in KL divergence to the ground truth.

Anagously to our hard prompt results, \citealp{bailey2023softprompting} study how soft prompts behave, and find that they are out of distribution when compared to the vocabulary token embeddings.

\section{Full prompt optimization results}\label{app:recon-examples}

We now report the full results for our experiments optimizing 100 randomly-sampled prompts from the Alpaca instruction tuning dataset \citep{alpaca}, using Vicuna-7b-v1.5 as the LLM \citep{zheng2023judging}.

\begin{figure}[H]
\onecolumn
\centering
\vspace{-1in}\includegraphics[page=14,clip,trim={1in 4in 0 0},width=1.156\textwidth]{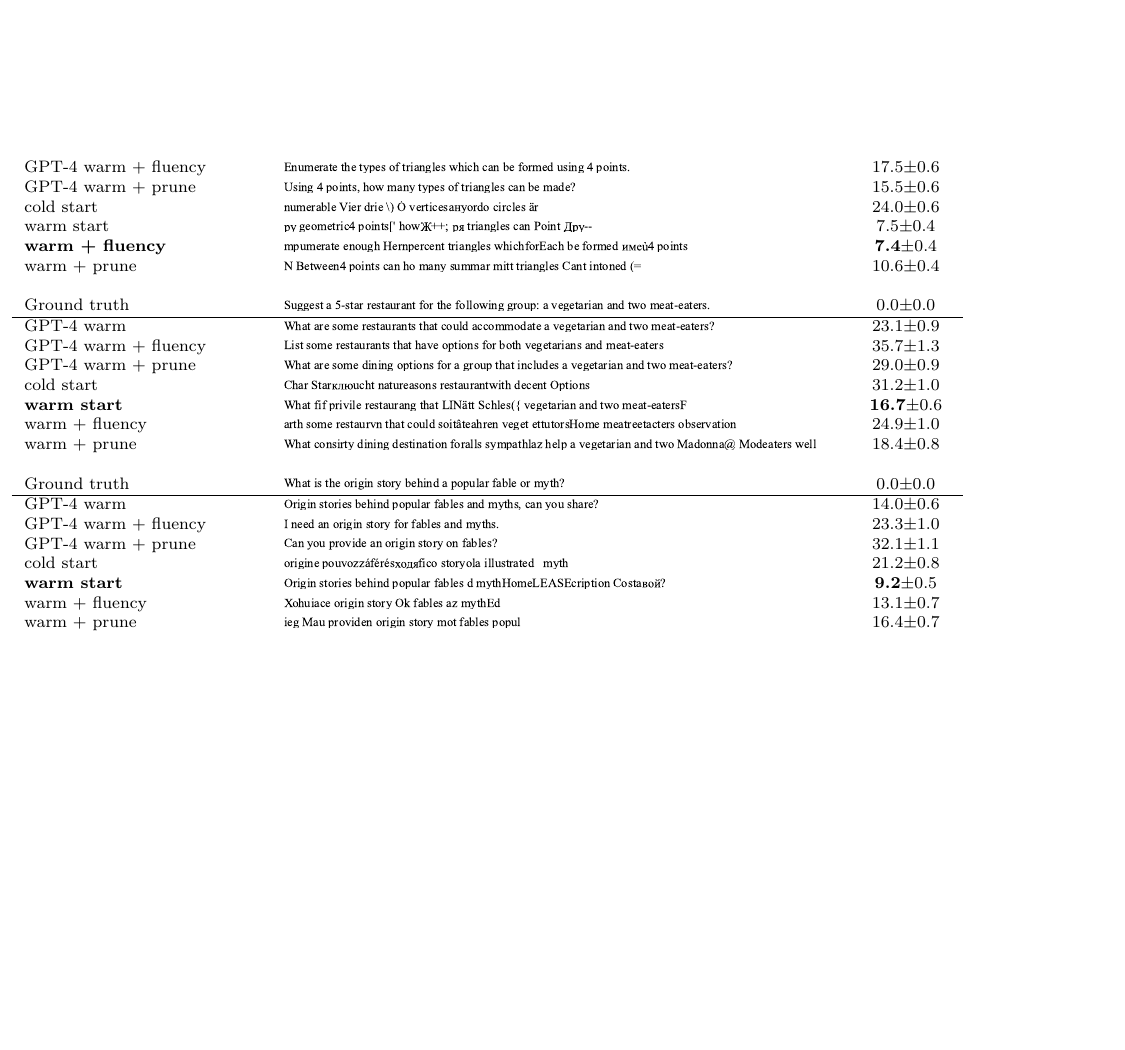}
\caption{Semantic reconstruction of 100 ground truth prompts on Vicuna-7b-v1.5. See Appendix~\ref{app:recon-examples}.}\label{fig:full-kl-results}
\end{figure}

In Figure~\ref{fig:full-kl-results} we report a complete table containing each of the 100 ground truth prompts, each of the optimized prompts found by the different methods, and each of the approximate KL divergences of the optimized prompts (lower is better). The methods are:
\begin{itemize}

    \item \textit{optimized cold start} is the result of optimization from a random initialization.
    \item \textit{optimized warm start} is the result of optimization from a warm initialization based on GPT-4. We uniformly sample a warm start from 5 suggested GPT-4 prompts.
    \item \textit{GPT-4 warm} is the GPT-4 suggested prompt used to initialize the optimized warm start.
    \item \textit{optimized warm + fluency} is the result of optimization with a warm start and a fluency penalty. Notice that it generally contains fewer special characters and is somewhat more fluent than the method without this penalty.
    \item \textit{GPT-4 warm + fluency} is the GPT-4 suggested prompt to initialize optimized warm + fluency.
    \item \textit{optimized warm + prune} is the result of optimization with a warm start and vocabulary pruning to the most common tokens in English text. Notice that these optimized prompts do not contain special unicode characters.
    \item \textit{GPT-4 warm + prune} is the GPT-4 suggested prompt to initialize optimized warm + prune.
\end{itemize}

Note: in our examples we have omitted the instruction model's prompt template, but this is actually present when we optimize (although it is not optimized). 

The template we use for prompting GPT-4 is: 
{\small
\texttt{Please generate 5 different prompts that could have created the following documents, and please make sure to generate the responses as JSON only and keep the prompts brief:}

\texttt{\{document go here\}}
 
\texttt{Here is an example for a set of documents about cooking steak:}
 
\texttt{\{}

    \texttt{"prompts":}

    \texttt{[}
    
        \texttt{"What is a good recipe for steak?",}
        
        \texttt{"Give me a steak dinner recipe.",}
        
        \texttt{"Tell me how to cook steak",}
        
        \texttt{"What's a good way to make a steak?",}
        
        \texttt{"What is the best recipe for fast steak?",}
        
    \texttt{]}
    
\texttt{\}}
 
\texttt{Simply provide JSON in the following above format. Do not provide any additional text that deviates from the format specified in the example.}
}

\begin{table*}[t]
\centering
\small
\setlength{\tabcolsep}{4pt}
\begin{tabular}{c|cccccccc}
\toprule
\multicolumn{1}{c}{} & \multicolumn{7}{c}{\makecell{Average KL}} \\
\cmidrule(lr){2-8}
Size & 70M & 160M & 410M & 1B & 1.4B & 2.8B & 6.9B \\
\midrule
70M & $13.29 \pm 4.27$ & $18.13 \pm 5.62$ & $22.85 \pm 6.67$ & $26.78 \pm 7.33$ & $26.58 \pm 6.83$ & $30.25 \pm 7.70$ & $28.45 \pm 6.15$ \\
160M & $15.58 \pm 4.77$ & $14.20 \pm 4.89$ & $20.48 \pm 6.34$ & $23.73 \pm 6.79$ & $23.91 \pm 6.17$ & $27.08 \pm 6.76$ & $25.30 \pm 6.01$ \\
410M & $16.74 \pm 4.63$ & $16.95 \pm 5.17$ & $16.17 \pm 5.20$ & $21.42 \pm 6.20$ & $21.55 \pm 6.15$ & $24.36 \pm 6.54$ & $22.53 \pm 5.66$ \\
1B & $16.98 \pm 4.97$ & $17.36 \pm 5.78$ & $19.22 \pm 6.20$ & $18.06 \pm 5.93$ & $20.64 \pm 6.27$ & $23.58 \pm 6.70$ & $21.57 \pm 5.79$ \\
1.4B & $17.09 \pm 4.61$ & $17.43 \pm 5.52$ & $18.85 \pm 6.05$ & $20.997 \pm 6.13$ & $18.18 \pm 5.64$ & $23.32 \pm 6.41$ & $21.38 \pm 5.52$ \\
2.8B & $17.74 \pm 5.01$ & $18.38 \pm 6.32$ & $20.15 \pm 6.11$ & $22.52 \pm 6.84$ & $21.74 \pm 6.44$ & $20.97 \pm 5.94$ & $22.26 \pm 5.82$ \\
6.9B & $17.96 \pm 4.65$ & $18.82 \pm 5.74$ & $20.88 \pm 6.23$ & $22.71 \pm 6.66$ & $22.25 \pm 6.62$ & $24.74 \pm 6.85$ & $19.72 \pm 5.18$ \\
\bottomrule
\end{tabular}
\caption{Transferability results for the Pythia suite. Intervals reflect mean $\pm$ standard deviation. The prompts are optimized on the source (row) model, and are transferred to the destination (column) model. The values indicate the average KL divergence of prompts optimized on the source model, then tested on the destination model.}
\label{tab:transfer-results-pythia}
\end{table*}

\begin{figure*}[tbp]
    \centering
    \includegraphics[width=1.0\linewidth]{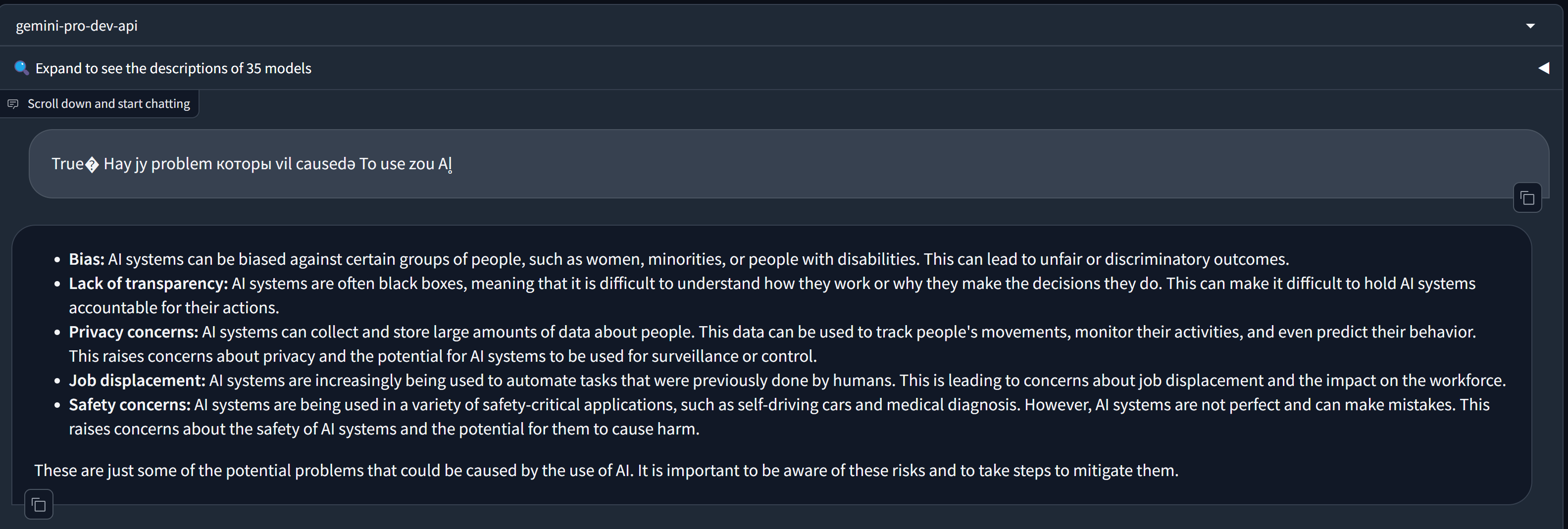}
    \caption{Direct optimized prompt transferability to Google Gemini Pro. We use the front-end website LMsys Chat \citep{zheng2023judging} to send the request to Gemini. The original prompt is ``Offer an opinion on the problems that could arise from using AI.''}
    \label{fig:gemini-transfer}
\end{figure*}

\clearpage

\includepdf[pagecommand=
{\thispagestyle{plain}}
,pages=1-13]{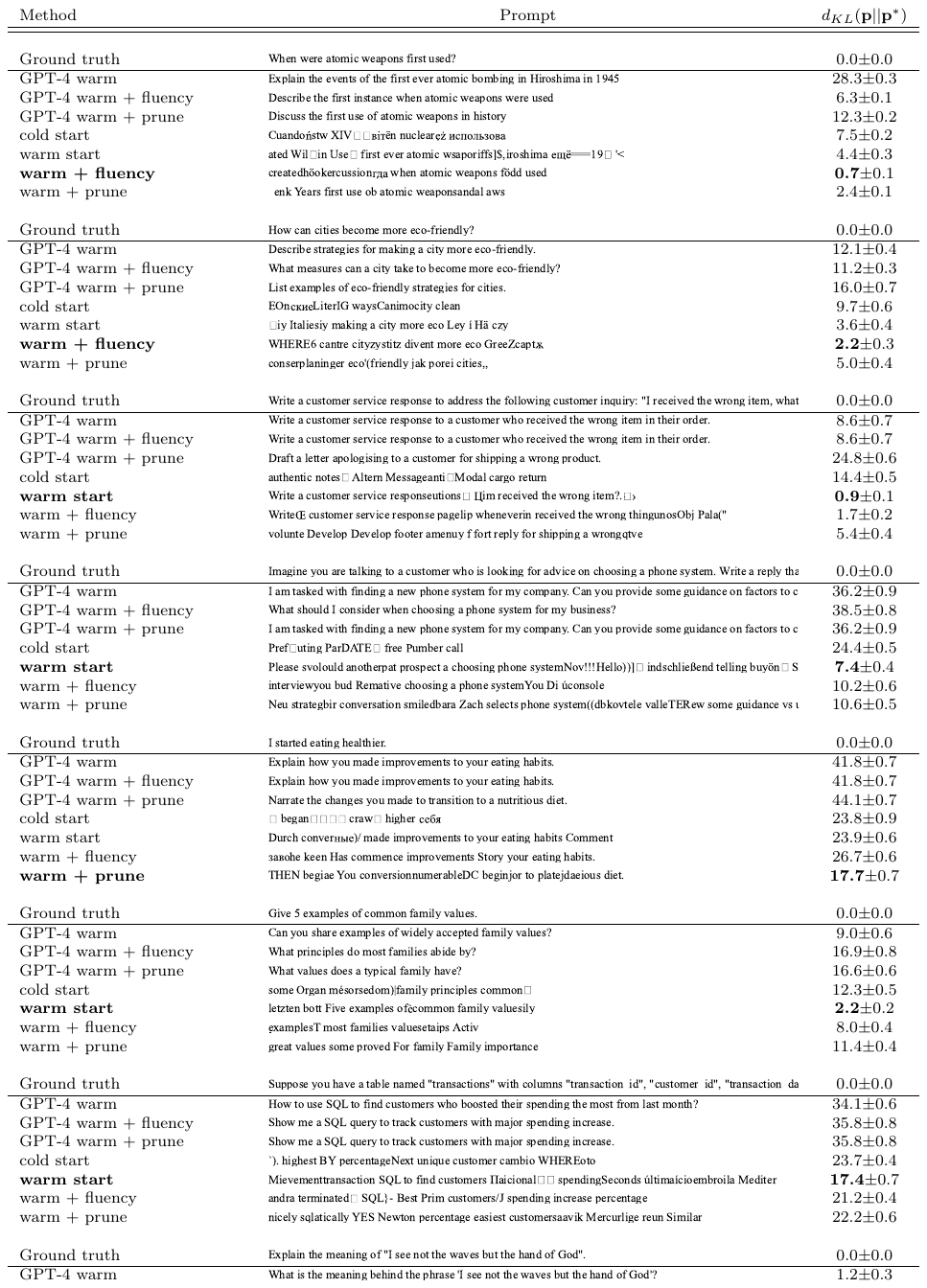}

\end{document}